\documentclass[runningheads]{llncs}

\usepackage[T1]{fontenc}
\usepackage{graphicx}
\graphicspath{{figs/}{zero_shot_llm_comparison/figs/}}
\usepackage{amsmath}
\usepackage{amssymb}
\usepackage{xcolor}
\usepackage[colorlinks=true,allcolors=blue]{hyperref}
\usepackage{subfig}
\usepackage[capitalise]{cleveref}
\crefname{figure}{Fig.}{Figs.}
\Crefname{figure}{Fig.}{Figs.}
\crefname{subfigure}{Fig.}{Figs.}
\Crefname{subfigure}{Fig.}{Figs.}
\usepackage{algorithm}
\usepackage[noend]{algorithmic}

\usepackage{booktabs}
\usepackage{needspace}
\usepackage{array}
\usepackage{adjustbox}
\usepackage{float}
\usepackage{fvextra}
\usepackage{tcolorbox}
\tcbuselibrary{skins,breakable}

\definecolor{promptframe}{RGB}{118,135,158}
\definecolor{promptfill}{RGB}{247,248,251}

\title{Benchmarking Zero-Shot LLM-Generated Parent Selection in Genetic Programming for Symbolic Regression}
\titlerunning{Benchmarking Zero-Shot LLM Parent Selection in GP}
\author{Hengzhe Zhang\inst{1} \and Qi Chen\inst{1} \and Bing Xue\inst{1} \and Wolfgang Banzhaf\inst{2} \and Mengjie Zhang\inst{1}}
\authorrunning{H. Zhang et al.}
\institute{Centre for Data Science and Artificial Intelligence \& School of Engineering and Computer Science, Victoria University of Wellington, PO Box 600, Wellington 6140, New Zealand\\
\email{\{hengzhe.zhang,qi.chen,bing.xue,mengjie.zhang\}@ecs.vuw.ac.nz}
\and
Department of Computer Science and Engineering, Michigan State University, East Lansing, MI 48824, USA\\
\email{banzhafw@msu.edu}}

\begin{document}

    \maketitle
    \setcounter{footnote}{0}

    \begin{abstract}
        Parent selection significantly affects exploration, exploitation, and complexity control in genetic programming (GP) for symbolic regression. It is unclear whether large language models (LLMs) can synthesize effective operators in a zero-shot setting without iterative meta-evolution. Here, zero-shot means that the model receives only the task description, with no reference operators or iterative feedback. In this work, we benchmark zero-shot synthesis of parent-selection operators across eight LLMs within a standard GP framework for symbolic regression. Each model receives the same natural-language prompt to generate a parent-selection operator, which is then evaluated in a standard GP framework with only the parent-selection operator replaced, while all other components and the evolutionary-search budget are held constant. For each LLM, ten independent zero-shot operators are evaluated on twelve OpenML regression benchmarks and compared against automatic lexicase and tournament selection baselines. Claude Sonnet~4.6 and Gemini~3.1 Pro stand out for consistently strong performance on both training and held-out test $R^2$. The strongest operator in our benchmark---a Kimi~K2.5 zero-shot synthesis---surpasses the automatic lexicase and tournament baselines in search effectiveness. These results suggest that zero-shot LLM synthesis is a viable approach to generating competitive GP selection operators. Analysis shows that many generated operators use semantics to guide selection, suggesting that LLMs can produce non-trivial search heuristics from the task description alone. We also examine the relationship between public LLM leaderboard rankings and GP performance. Widely used benchmarks, such as Humanity's Last Exam and SWE-bench Verified, strongly correlate with training $R^2$, while their relationship to held-out test $R^2$ is weaker and less clear.
    \end{abstract}

    \keywords{genetic programming \and parent selection \and symbolic regression \and large language models \and zero-shot program synthesis \and evolutionary computation}

    \section{Introduction}
    Genetic programming (GP) is a widely used evolutionary approach for symbolic regression, where evolved program trees predict continuous outputs from input features~\cite{banzhaf1998genetic}. Parent selection is a central component in this process: it decides which programs are selected as parents to generate offspring and influences the exploration--exploitation trade-off~\cite{xie2012parent}. Parent selection also interacts with parsimony pressure and bloat~\cite{luke2002lexicographic}. While hand-crafted operators such as tournament selection~\cite{banzhaf1998genetic} and lexicase selection~\cite{la2019probabilistic} are well-studied, designing effective selection operators remains an open challenge that motivates ongoing research~\cite{zhang2025llm}.

    Large language models (LLMs) can now synthesize executable code from natural language, opening a new route to automated algorithm design for evolutionary algorithm (EA) workflows~\cite{zhang2025llm}. This potential has already been realized in systems such as LLaMEA~\cite{van2024llamea}, which uses LLMs to generate metaheuristics. More broadly, recent work has explored LLM-enhanced EAs across neural architecture search~\cite{lai2026llmenas}, scheduling heuristic design~\cite{yu2026automated}, hybrid flow-shop scheduling~\cite{zhao2026large}, and autonomous multi-objective optimization~\cite{huang2025autonomous}. Much of the reported success, however, comes from multi-round systems rather than zero-shot synthesis from a single prompt. FunSearch~\cite{romera2024mathematical} repeatedly improves programs through archive-guided iteration. Evolution of Heuristics~\cite{liu2024evolution} evolves both heuristic descriptions and their implementations. ReEvo~\cite{ye2024reevo} adds reflection signals to guide subsequent generations. These approaches introduce an additional outer-loop evolutionary process that can compensate for weak individual generations, making it difficult to isolate the intrinsic capability of LLMs in zero-shot code synthesis. At the same time, LLM-driven code generation still leans heavily on a few flagship systems---most visibly GPT-family models---whereas systematic head-to-head comparisons across a broad range of contemporary LLMs on the same algorithm-design task remain limited. Therefore, zero-shot LLM synthesis for algorithm design remains poorly understood.

    This paper investigates whether LLMs can produce effective parent-selection operators in a zero-shot synthesis setting. Within a standard GP framework, we vary only the selection stage while keeping variation and fitness evaluation fixed, ensuring that any observed differences can be attributed to selection behavior. The goal is to characterize zero-shot synthesis capability across LLMs, determine which patterns emerge from those intrinsic capabilities, and identify which public LLM leaderboard scores can predict the performance of generated operators. Overall, the paper pursues three objectives:
    \begin{itemize}
        \item To establish a benchmark for evaluating LLMs' zero-shot ability to produce parent-selection operators for symbolic-regression GP within a standard GP framework and to characterize the capabilities of a broad range of modern LLMs on this algorithm-design task.\footnote{The generated operators are available at \url{https://github.com/hengzhe-zhang/Zero-Shot-LLM-Benchmark}.}
        \item To identify the algorithmic patterns that emerge in LLM-designed selection operators without prior knowledge.
        \item To assess which public LLM leaderboard scores align most closely in rank order with GP-task performance, offering practical guidance for model selection prior to domain-specific evaluations.
    \end{itemize}

    The remainder of this paper is organized as follows. \Cref{sec:related-work} summarizes related research on LLM-based synthesis and GP selection. \Cref{sec:benchmark-protocol} presents the benchmark design of zero-shot LLM-generated selection for GP, and \Cref{sec:experimental-setup} details the experimental setup. \Cref{sec:results-analysis} reports empirical findings on zero-shot performance, recurring structure in generated operators, and alignment with LLM leaderboards. \Cref{sec:conclusion} concludes the paper and outlines future directions.

    \section{Related Work}\label{sec:related-work}

    \subsection{LLMs for Evolutionary Algorithm Design}
    LLMs have been increasingly applied to automated algorithm design within evolutionary algorithms~\cite{wu2024evolutionary}, including code evolution~\cite{hemberg2024evolving}, evolutionary neural architecture search~\cite{lai2026llmenas}, scheduling heuristic generation~\cite{yu2026automated}, hybrid flow-shop scheduling~\cite{zhao2026large}, and autonomous multi-objective optimization~\cite{huang2025autonomous}. LLaMEA~\cite{van2024llamea} evolves candidate metaheuristic code with performance-based selection. FunSearch~\cite{romera2024mathematical} repeatedly prompts the model with high-scoring programs from an archive, evaluates generated code, and inserts validated improvements back into the archive. Evolution of Heuristics~\cite{liu2024evolution} uses evolutionary operators over both heuristic descriptions and their code implementations. ReEvo~\cite{ye2024reevo} combines evolutionary search with reflection signals from the LLM to guide subsequent candidate generation. LLM-assisted metaheuristic evolution has also been explored for swarm algorithms such as the Self-Organizing Migrating Algorithm (SOMA)~\cite{pluhacek2024using}. These procedures are multi-round pipelines that mix one-shot code quality with search pressure from the outer evolutionary loop, obscuring the intrinsic generation capability of a single model call. Early work suggests that an LLM without iterative evolution struggles to design good selection heuristics, but that evidence targets older models such as GPT-4~\cite{zhang2024understanding}. This raises the question of how well modern LLMs perform in zero-shot conditions.

    \subsection{Selection in Genetic Programming}
    Selection strongly influences GP search trajectories. Classical GP work has long studied standard tournament selection and fitness-proportionate selection~\cite{banzhaf1998genetic}, while later work explored dynamically adjusting tournament selection pressure~\cite{xie2012parent}. Multi-objective GP has also explored Pareto-based tournament variants to balance competing goals such as predictive accuracy and model complexity~\cite{kotanchek2006pursuing}. Lexicase selection was introduced for program synthesis via case-wise filtering~\cite{helmuth2014solving}. Subsequent work developed several lexicase-style variants for regression, including continuous relaxations for non-discrete fitness~\cite{la2019probabilistic}, probabilistic speedups~\cite{ding2023probabilistic}, batch tournament selection for lexicase-like quality at tournament-like speed~\cite{de2019batch}, minimum-variance thresholding for more principled per-filter thresholds~\cite{imai2024minimum}, and DALex as an efficient approximation that retains effectiveness~\cite{ni2024dalex}. Parsimony-aware selection has been studied through lexicographic parsimony pressure~\cite{luke2002lexicographic} and extended to lexicase selection~\cite{de2022lexi2}. Diversity-aware selection has been studied via novelty search~\cite{lehman2010efficiently} and complementary-phenotype selection for crossover~\cite{dolin2002opposites}. With the rise of LLMs for automatic algorithm design, it is natural to ask how well this technique performs when used for GP and whether its synthesis capability varies across models.

    \section{Benchmark Design}\label{sec:benchmark-protocol}

    \subsection{Overview of the Benchmark Design}

    As illustrated in \Cref{fig:benchmark-loop}, the overall pipeline begins with a fixed natural-language prompt, from which each LLM generates a selection operator as executable code. To isolate the effect of parent selection and enable a controlled comparison, these operators are evaluated within a common GP system where all other components are held constant. Each generated operator is tested over a fixed benchmark collection $\mathcal{B}$ of twelve OpenML regression datasets, which is a representative subset of SRBench~\cite{imai2025call}. For every dataset $\mathcal{D}\in\mathcal{B}$, the standard GP process is run independently, and performance is measured by training and test $R^2$.

    \subsection{Evolutionary Framework}
    \begin{figure}[!t]
        \centering
        \includegraphics[width=\linewidth,trim=5pt 5pt 5pt 5pt,clip]{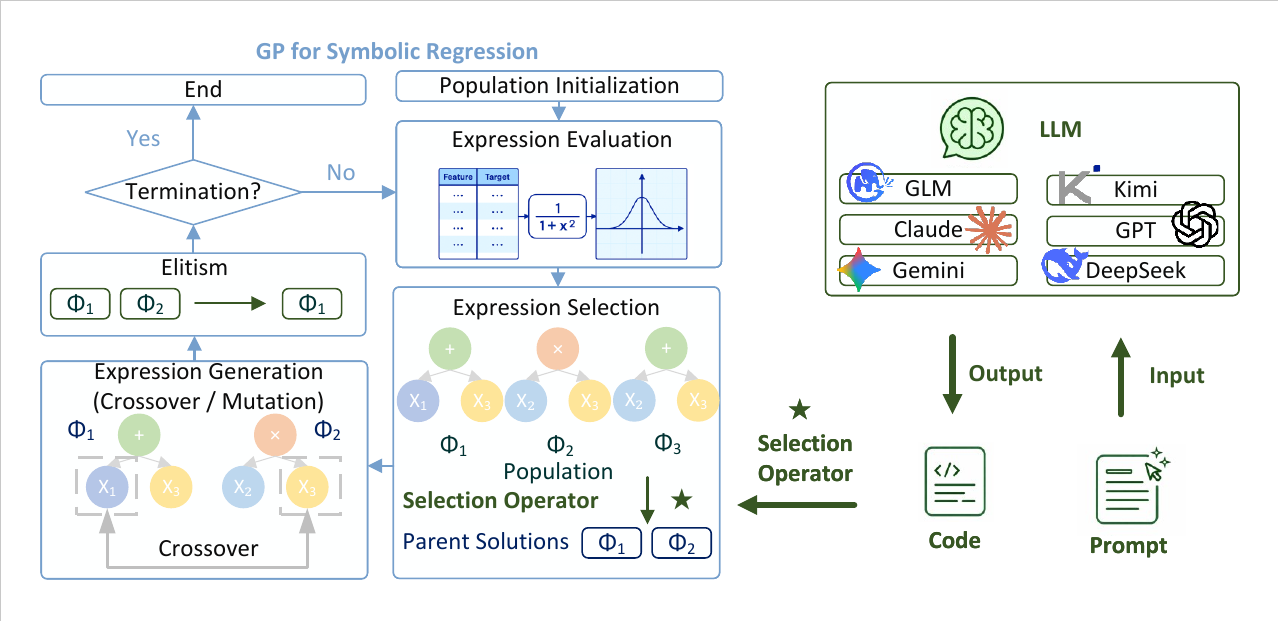}
        \caption{Benchmark evolutionary loop for GP-based symbolic regression. Only the parent-selection operator is replaced by LLM-generated code, while all other components and budgets are the same.}
        \label{fig:benchmark-loop}
    \end{figure}

    Each operator is embedded in a standard GP framework for symbolic regression. For any dataset $\mathcal{D}=\{(\mathbf{x}_i,y_i)\}_{i=1}^{m}$, symbolic regression searches for an expression with strong predictive quality. During a run on $\mathcal{D}$, each individual, represented as a program tree, produces predictions $\hat{y}_i$ and per-case errors $e_i$ on the fitness cases, and these values are provided to the parent-selection operator. Let $\mathcal{P}_g$ denote the GP population at generation $g$, and let $S$ denote the parent-selection operator, which returns $|\mathcal{P}_g|$ selected individuals from $\mathcal{P}_g$. For each generated operator, quality is measured by per-dataset training and test $R^2$ after running GP-based symbolic regression with the same crossover, mutation, and fitness-evaluation procedure. Specifically:
    \begin{itemize}
        \item \textbf{Initialization:} The initial population is generated using the ramped half-and-half method to obtain a diverse set of tree depths and structures at generation $g=0$~\cite{banzhaf1998genetic}.
        \item \textbf{Selection:} Parent selection is the only component replaced by LLM-generated code. Given $\mathcal{P}_g$, the generated operator $S$ returns a parent set of size $|\mathcal{P}_g|$, which is then used for crossover and mutation.
        \item \textbf{Variation:} Offspring are produced using standard subtree crossover and subtree mutation~\cite{banzhaf1998genetic}. Crossover exchanges randomly chosen subtrees between selected parents, while mutation replaces a randomly chosen subtree with a newly sampled subtree.
        \item \textbf{Fitness evaluation:} Each individual is evaluated on the training set, with linear scaling of predictions and $R^2$ as the performance criterion. Case-wise errors on fitness cases are computed for every individual and provided to the parent-selection operator to support diversity maintenance.
        \item \textbf{Elitism:} A top-1 elitist strategy preserves the best individual from the current generation, so the best-so-far solution is kept across generations.
    \end{itemize}
    The left panel of \Cref{fig:benchmark-loop} shows the evolutionary process.

    \subsection{Zero-Shot Operator Generation}
    Selection operators are synthesized in a zero-shot manner: each LLM must generate the operator from the task description alone, with no reference operators, iterative feedback, or meta-level evolutionary loop that recombines or mutates previously generated operator code. Each LLM returns executable code that, after parsing, undergoes a lightweight evaluation pass before any benchmark GP run. This lightweight evaluation checks whether the code is syntactically correct and executable, and runs several basic functional test cases to verify that the selection operator returns the required number of individuals. It does not involve any symbolic-regression evaluation. Operators that raise runtime errors, fail these test cases, or exceed a time limit are discarded and regenerated until 10 valid operators have been obtained. The natural-language prompt is identical across LLMs and does not encode expert knowledge beyond the task description. The full prompt is shown in \Cref{fig:zero-shot-prompt}.

    \begin{figure}[!t]
        \centering
        \begin{tcolorbox}[
            enhanced jigsaw,
            breakable,
            width=\linewidth,
            colback=promptfill,
            colframe=promptframe,
            boxrule=0.55pt,
            arc=1.1mm,
            left=9pt,
            right=9pt,
            top=5pt,
            bottom=5pt,
            boxsep=0pt,
            before skip=0pt,
            after skip=\smallskipamount,
        ]
        {\linespread{0.85}\selectfont
            \begin{Verbatim}[
                fontsize=\small,
                breaklines,
                breakindent=1.5em,
            ]
    When writing code, prefer NumPy vectorized operations and avoid explicit Python for-loops unless absolutely necessary. Please implement code within 30 lines.
    Your task is to develop an innovative and novel selection operator for symbolic regression using genetic programming in Python.
    Ensure that your newly designed function adheres to the following signature:
    def custom_selection(population, k=100, status={}):
        # Useful information about individuals:
        # squared_error_vector = individual.case_values
        # predicted_values = individual.predicted_values
        # residual = individual.y-individual.predicted_values
        # number_of_nodes = len(individual)
        # height = individual.height
        # Useful information about evolution:
        # status["evolutionary_stage"]: [0,1], where 0 is the first generation and 1 is the final generation
        # Implement selection logic here
        return selected_individuals
    You do not need to provide a usage example.
    Embrace creativity, novelty, and bold experimentation to push the boundaries of the state of the art in selection operators for genetic programming.
            \end{Verbatim}
        }
        \end{tcolorbox}
        \caption{Prompt used for zero-shot generation of GP parent-selection operators.}
        \label{fig:zero-shot-prompt}
    \end{figure}

    \section{Experimental Setup}\label{sec:experimental-setup}

    \subsection{Parameter Settings}

    The experimental design compares LLMs on zero-shot synthesis of parent-selection operators under an identical GP benchmarking setup. \Cref{tab:selection-parameter-settings} summarizes the key parameter settings used in our experiments. Categorical features are encoded using target encoding~\cite{micci2001preprocessing} fitted on the training data. The function set includes arithmetic operations and mathematical functions. The analytical quotient $\mathrm{AQ}(x, y) = x / \sqrt{1 + y^2}$ avoids division-by-zero errors; $\mathrm{AbsLog}(x) = \log(1 + |x|)$ keeps logarithmic transforms well-defined. $\text{Sin}_\pi(x)$ and $\text{Cos}_\pi(x)$ represent $\sin(\pi x)$ and $\cos(\pi x)$, respectively, since the constant $\pi$ is difficult to evolve.

    \begin{table}[!t]
        \centering
        \footnotesize
        \caption{Parameter settings used in the experiments.}
        \label{tab:selection-parameter-settings}
        \begin{tabular}{lc}
            \toprule
            \textbf{Parameter}    & \textbf{Setting} \\
            \midrule
            Population size       & 200              \\
            Number of generations & 100              \\
            Crossover probability & 0.9              \\
            Mutation probability  & 0.1              \\
            Function set &
            \begin{tabular}[t]{@{}l@{}}
                \texttt{Add, Sub, Mul, AQ, Sqrt, AbsLog, Abs, Square, Sin$_\pi$, Cos$_\pi$} \\
                \texttt{Max, Min, Neg}
            \end{tabular} \\
            \bottomrule
        \end{tabular}
    \end{table}

    \begin{figure}[!t]
        \centering
        \includegraphics[width=0.45\linewidth]{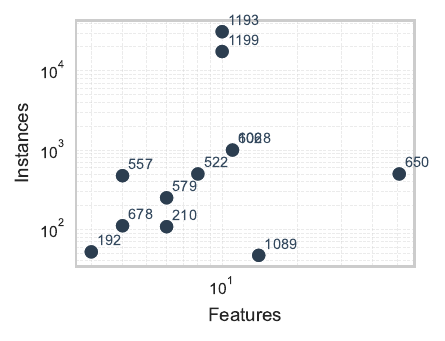}
        \caption{Twelve benchmark datasets: instance count versus feature count on log--log axes. Point labels are OpenML dataset identifiers.}
        \label{fig:benchmark-openml-dataset-sizes}
    \end{figure}

    \subsection{Datasets}
    We evaluate on twelve representative SRBench problems, which have been systematically selected through clustering~\cite{imai2025call}, with OpenML IDs 1028, 1089, 1193, 1199, 192, 210, 522, 557, 579, 606, 650, 678. \Cref{fig:benchmark-openml-dataset-sizes} shows the twelve datasets in the instance--feature plane on log--log scales. Instance counts range from 47 to 31{,}104, and the number of input features ranges from 3 to 51.

    \subsection{Evaluation Protocol}
    For each LLM, we generate ten zero-shot parent-selection operators. Each operator is then evaluated on all twelve benchmark datasets, and for each dataset we use an 80/20 train--test split with five different random seeds, yielding $12 \times 5 = 60$ GP runs per operator and $600$ runs per LLM. We use the coefficient of determination, $R^2 = 1 - \tfrac{\sum_i (y_i - \hat{y}_i)^2}{\sum_i (y_i - \bar{y})^2}$, computed on the train or test sets, as the evaluation metric.

    \section{Experimental Results}\label{sec:results-analysis}
    The empirical study is organized around the following research questions:
    \begin{itemize}
        \item \textbf{RQ1:} How well do the benchmarked LLMs perform at generating zero-shot parent-selection operators within a standard GP framework under a shared prompt?
        \item \textbf{RQ2:} What recurring patterns emerge in LLM-generated selection operators, and how do they differ across LLMs?
        \item \textbf{RQ3:} Which existing public leaderboard scores serve as useful rank-order surrogates for identifying suitable LLMs on selection-operator design tasks, prior to domain-specific synthesis runs?
    \end{itemize}

    \subsection{RQ1: Zero-Shot Performance}\label{subsec:rq1-zero-shot-performance}
    \begin{figure*}[!t]
        \centering
        \includegraphics[width=0.92\textwidth]{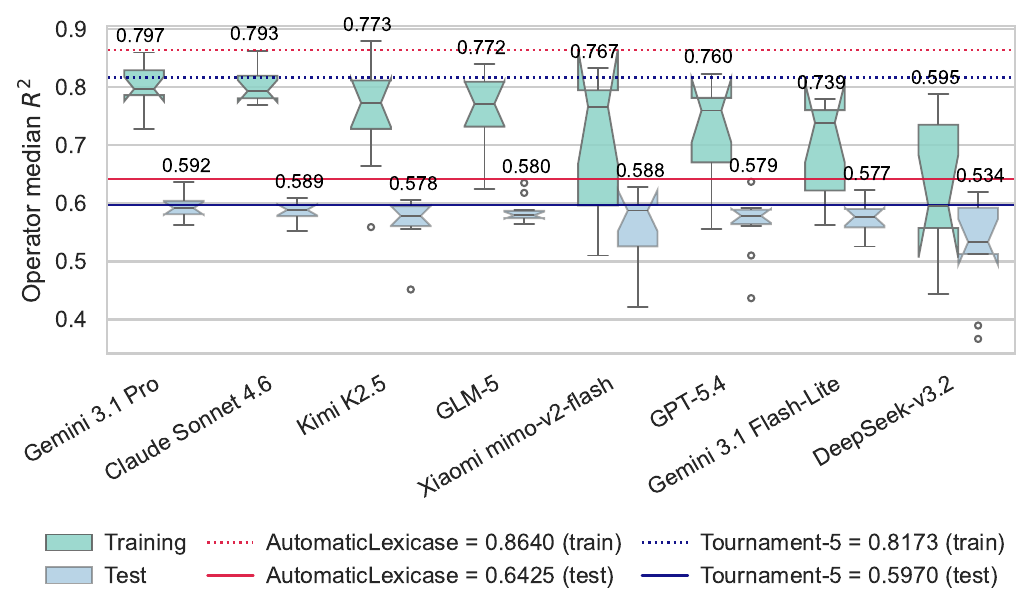}
        \caption{Box plots of operator scores for the ten zero-shot operators generated by each LLM. Each operator's score is summarized over all datasets and five split seeds.}
        \label{fig:zero-shot-boxplot}
    \end{figure*}

    \Cref{fig:zero-shot-boxplot} compares zero-shot LLM-generated selection operators in terms of training and test $R^2$. Each box plot summarizes ten operators per LLM, aggregated over all datasets and five random seeds. The dashed horizontal lines show the performance of automatic lexicase and tournament selection with tournament size 5, computed on the same datasets and under the same evolutionary-search budget as the LLM-generated operators.

    \begin{table}[!t]
        \centering
        \caption{Zero-shot \textbf{training} performance summary across benchmarked LLMs with $N=200$ and $G=100$. Parenthetical integers rank models within each metric column.}
        \resizebox{\columnwidth}{!}{%
            \begin{tabular}{c|c|c|c|c|c|c}
                \toprule
                Model/Operator        & Mean over mean      & Median over mean    & Mean over median    & Median over median & Best score & Mean rank \\
                \midrule
                Claude Sonnet 4.6     & \textbf{0.7019 (1)} & 0.6921 (2)          & \textbf{0.8036 (1)} & 0.7935 (2)          & 0.8633 (2) & \textbf{1.60} \\
                Gemini 3.1 Pro        & 0.7015 (2)          & \textbf{0.7003 (1)} & 0.8029 (2)          & \textbf{0.7972 (1)} & 0.8601 (3) & 1.80 \\
                Kimi K2.5             & 0.6844 (4)          & 0.6885 (4)          & 0.7554 (4)          & 0.7735 (3)          & \textbf{0.8798 (1)} & 3.20          \\
                GLM-5                 & 0.6884 (3)          & 0.6897 (3)          & 0.7631 (3)          & 0.7718 (4)          & 0.8397 (4)          & 3.40          \\
                GPT-5.4               & 0.6692 (5)          & 0.6845 (5)          & 0.7210 (5)          & 0.7602 (6)          & 0.8238 (6)          & 5.40          \\
                Xiaomi mimo-v2-flash  & 0.6662 (6)          & 0.6765 (6)          & 0.7070 (6)          & 0.7668 (5)          & 0.8329 (5)          & 5.60          \\
                Gemini 3.1 Flash-Lite & 0.6585 (7)          & 0.6674 (7)          & 0.6986 (7)          & 0.7392 (7)          & 0.7805 (8)          & 7.20          \\
                DeepSeek-v3.2         & 0.6228 (8)          & 0.6318 (8)          & 0.6268 (8)          & 0.5951 (8)          & 0.7883 (7)          & 7.80          \\
                \midrule
                AutomaticLexicase & \multicolumn{2}{c|}{0.7298} & \multicolumn{2}{c|}{0.8640} & - & - \\
                Tournament-5 & \multicolumn{2}{c|}{0.7284} & \multicolumn{2}{c|}{0.8173} & - & - \\
                \bottomrule
            \end{tabular}
        }
        \label{tab:zero-shot-patterns}
        \vspace{-3mm}
    \end{table}

    \paragraph{Training $R^2$.}
    Training $R^2$ shows how effectively each generated selection operator improves search performance. \Cref{fig:zero-shot-boxplot} and \Cref{tab:zero-shot-patterns} summarize training $R^2$ results. For each of the ten zero-shot operators per LLM, we record training $R^2$ over twelve datasets and five train--test split seeds, for a total of $60$ runs. \Cref{tab:zero-shot-patterns} first computes each operator's mean and median training $R^2$ over its 60 runs, then aggregates these per-operator scores across the ten operators. Here, ``Mean over mean'' and ``Median over mean'' are the mean and median of the ten per-operator mean scores, whereas ``Mean over median'' and ``Median over median'' are the mean and median of the ten per-operator median scores. ``Best score'' denotes the largest median score. Mean-based scores reflect overall performance, whereas median-based scores are more robust to outliers.

\begin{figure*}[!t]
        \centering
        \subfloat[Training $R^2$.]%
        {\includegraphics[width=0.49\textwidth]{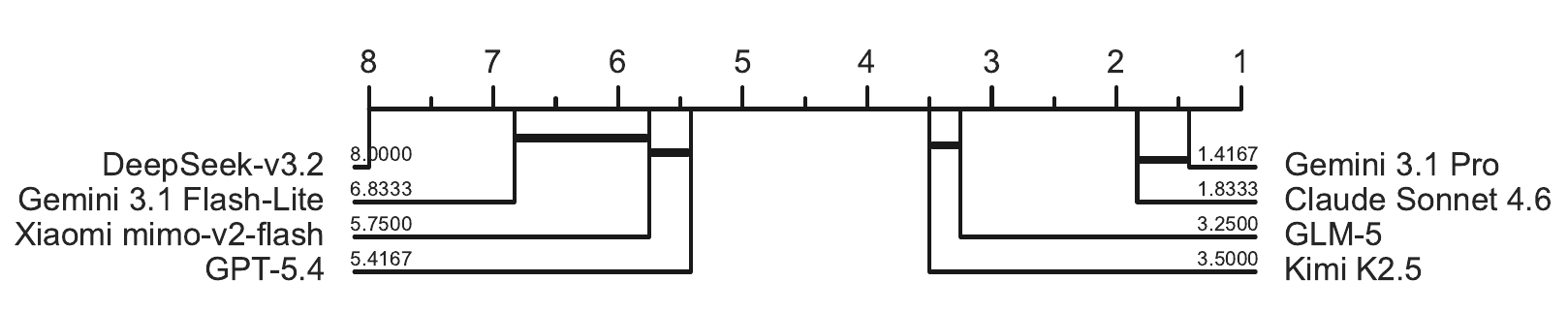}}
        \hfill
        \subfloat[Test $R^2$.]%
        {\includegraphics[width=0.49\textwidth]{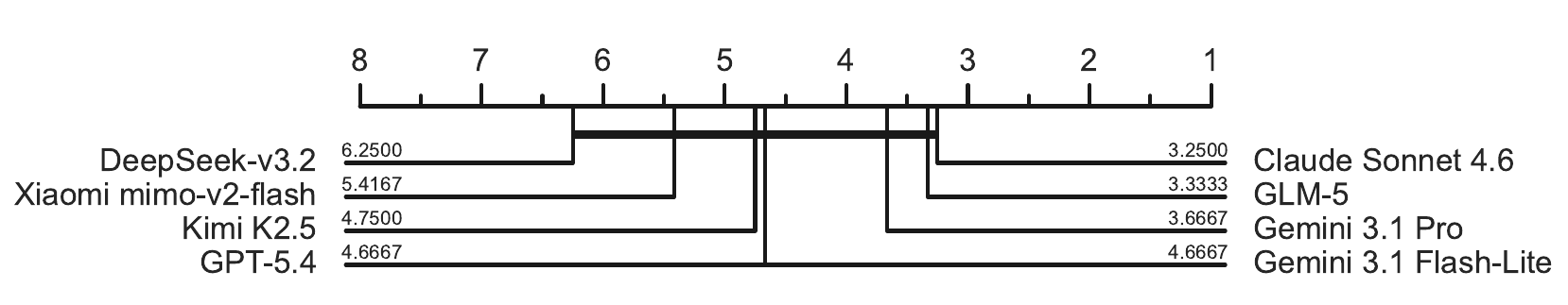}}
        \caption{Critical difference diagrams: Friedman ranks with Wilcoxon--Holm post-hoc correction on twelve datasets. Connected methods are not significantly different at $\alpha=0.05$. Lower rank is better.}
        \label{fig:zero-shot-cd-diagrams}
    \end{figure*}

    On training, Claude Sonnet~4.6 ranks first on mean-based summaries and achieves the best mean rank ($1.60$), Gemini~3.1 Pro ranks first on median-based summaries, and Kimi~K2.5 leads the best-score column (\Cref{tab:zero-shot-patterns}). Overall, Claude Sonnet~4.6 and Gemini~3.1 Pro show stronger typical performance across the ten zero-shot samples, whereas Kimi's best operator achieves the highest median score but its operators are weaker on average. As shown in \Cref{fig:zero-shot-boxplot}, the best operator produced by Kimi~K2.5 outperforms expert-designed operators in search effectiveness. \Cref{fig:zero-shot-cd-diagrams} shows critical difference diagrams. On training, Claude Sonnet~4.6 and Gemini~3.1 Pro are significantly better than the lower-ranked group.

    \begin{table}[!t]
        \centering
        \caption{Zero-shot \textbf{test} performance summary across benchmarked LLMs with $N=200$ and $G=100$. Parenthetical integers rank models within each metric column.}
        \resizebox{\columnwidth}{!}{%
            \begin{tabular}{c|c|c|c|c|c|c}
                \toprule
                Model/Operator        & Mean over mean      & Median over mean    & Mean over median    & Median over median & Best score & Mean rank \\
                \midrule
                Gemini 3.1 Pro        & 0.5415 (3)          & 0.5478 (3)          & \textbf{0.5944 (1)} & \textbf{0.5918 (1)} & 0.6370 (2) & \textbf{2.00} \\
                Claude Sonnet 4.6     & \textbf{0.5514 (1)} & \textbf{0.5607 (1)} & 0.5875 (2)          & 0.5890 (2)          & 0.6094 (7) & 2.60 \\
                GLM-5                 & 0.5491 (2)          & 0.5459 (5)          & 0.5865 (3)          & 0.5796 (4)          & 0.6353 (3)          & 3.40          \\
                Gemini 3.1 Flash-Lite & 0.5415 (3)          & 0.5482 (2)          & 0.5733 (4)          & 0.5773 (7)          & 0.6230 (5)          & 4.20          \\
                GPT-5.4               & 0.5347 (6)          & 0.5451 (6)          & 0.5644 (6)          & 0.5786 (5)          & \textbf{0.6373 (1)} & 4.80          \\
                Xiaomi mimo-v2-flash  & 0.5285 (7)          & 0.5465 (4)          & 0.5574 (7)          & 0.5877 (3)          & 0.6277 (4)          & 5.00          \\
                Kimi K2.5             & 0.5364 (5)          & 0.5414 (7)          & 0.5684 (5)          & 0.5785 (6)          & 0.6057 (8)          & 6.20          \\
                DeepSeek-v3.2         & 0.4927 (8)          & 0.5128 (8)          & 0.5252 (8)          & 0.5336 (8)          & 0.6200 (6)          & 7.60          \\
                \midrule
                AutomaticLexicase & \multicolumn{2}{c|}{0.5719} & \multicolumn{2}{c|}{0.6425} & - & - \\
                Tournament-5 & \multicolumn{2}{c|}{0.5645} & \multicolumn{2}{c|}{0.5970} & - & - \\
                \bottomrule
            \end{tabular}
        }
        \label{tab:zero-shot-test-summary}
        \vspace{-3mm}
    \end{table}

    \paragraph{Test $R^2$.}
    \Cref{fig:zero-shot-boxplot} and \Cref{tab:zero-shot-test-summary} apply the same aggregation scheme to held-out test $R^2$. As shown in \Cref{tab:zero-shot-test-summary}, Claude Sonnet~4.6 ranks first on mean-over-mean and median-over-mean, Gemini~3.1 Pro ranks first on mean-over-median, median-over-median, and mean rank, and GPT-5.4 leads the best-score column. Relative to training, several models move up or down across columns---for example, Kimi~K2.5 drops in several rankings despite leading the training best-score column. Compared to expert-designed operators, the zero-shot-generated selection operators are less competitive on the test set. Test performance is influenced by factors such as the function set and fitness-function design, which affect whether gains during training generalize to held-out data. Thus, some operators improve training performance but still overfit and underperform expert-designed operators on the test fold.

\begin{figure*}[!t]
        \centering
        \includegraphics[width=0.88\textwidth]{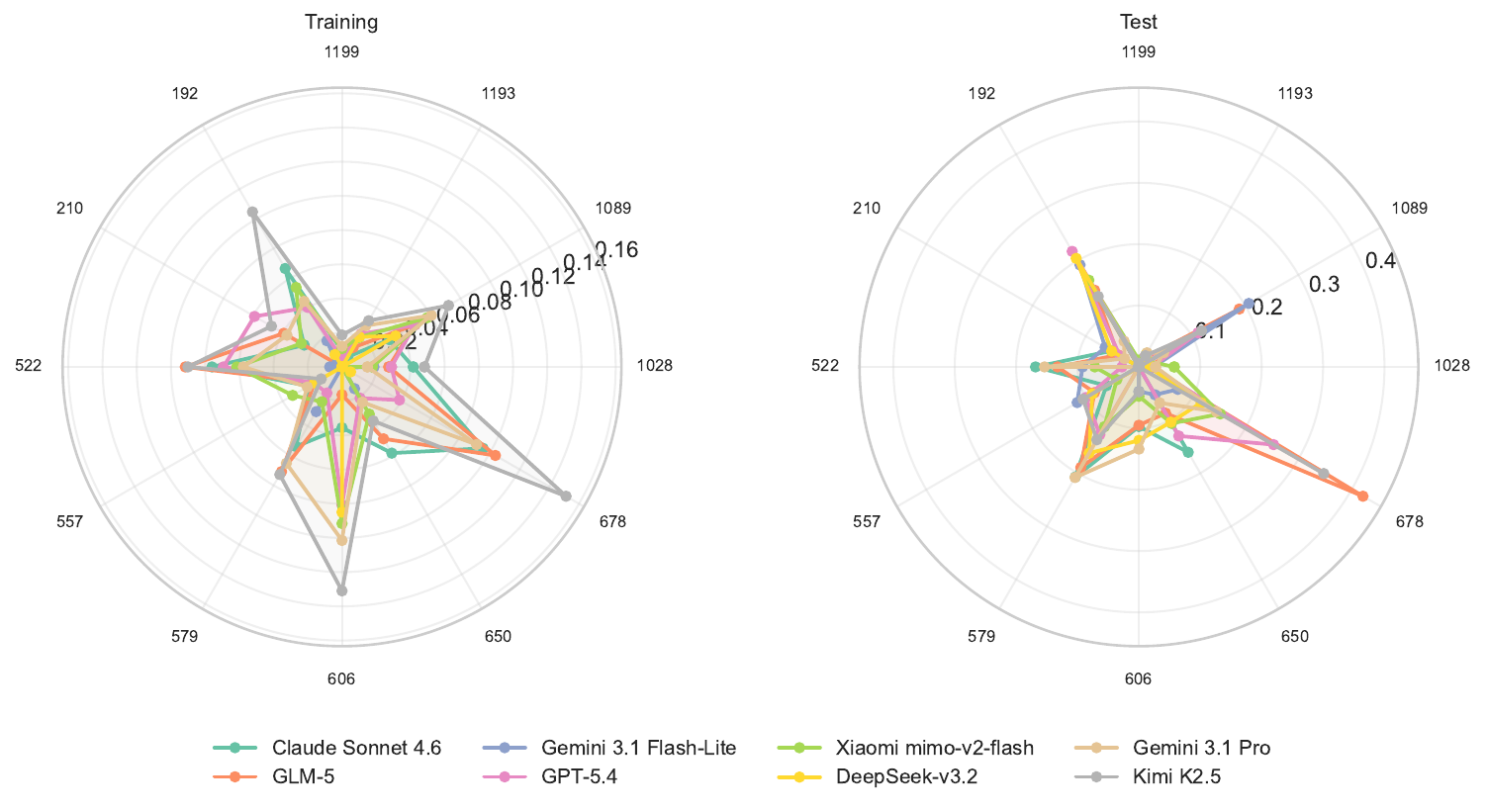}
        \caption{Per-dataset $R^2$ profiles after per-dimension minimum centering. \textbf{Left:} training-best operator per model; \textbf{right:} test-best operator per model.}
        \label{fig:zero-shot-best-operator-diamond}
    \end{figure*}

    \paragraph{Training versus test $R^2$.}
    Strong training performance does not always transfer to the test folds. \Cref{fig:zero-shot-best-operator-diamond} therefore shows the training and test $R^2$ scores of the best-performing operator on each dataset. Kimi~K2.5 exhibits strong training performance on datasets such as OpenML 606 but is substantially weaker on the corresponding test set, with the train--test gap varying by dataset. Even so, Claude Sonnet~4.6 and Gemini~3.1 Pro remain consistently competitive on both training and test $R^2$.

    \begin{figure*}[!t]
        \centering
        \subfloat[Distribution of operator running time.]%
        {\includegraphics[width=0.49\textwidth]{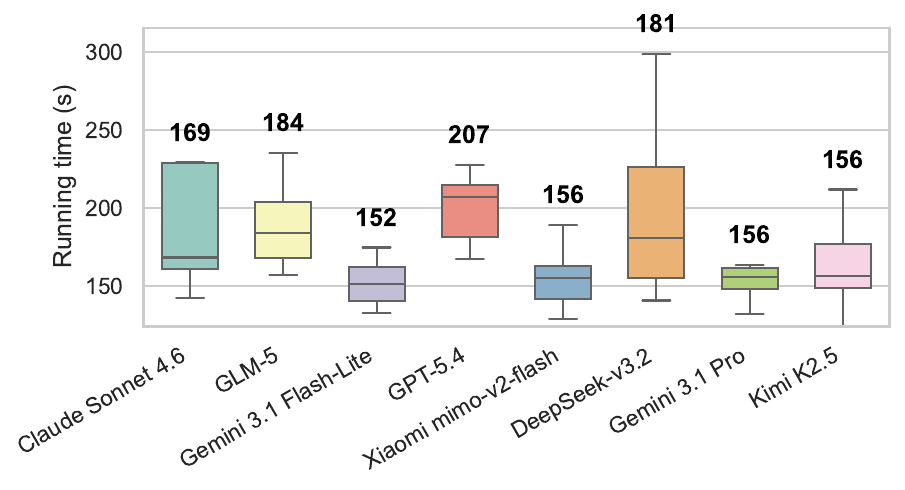}}
        \hfill
        \subfloat[Distribution of generated code length.]%
        {\includegraphics[width=0.49\textwidth]{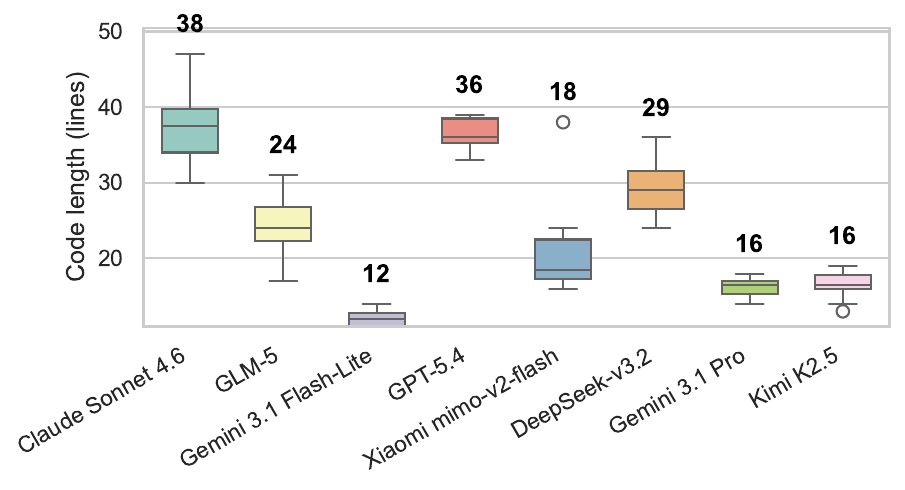}}
        \caption{Distributional comparison of LLM-generated operators across models. Left: running time (seconds); right: code length.}
        \label{fig:zero-shot-runtime-code-length}
    \end{figure*}

    \paragraph{Running time and code length.}
    \Cref{fig:zero-shot-runtime-code-length} plots the total GP runtime when using each generated operator and the line count of the generated operator code. Regarding running time, generated operators incur comparable computational cost, with no model exhibiting a consistently extreme runtime. In terms of code length, Kimi and Gemini tend to produce shorter operators, whereas the other LLMs tend to produce longer code.

    \begin{figure*}[!t]
        \centering
        \includegraphics[width=0.92\textwidth]{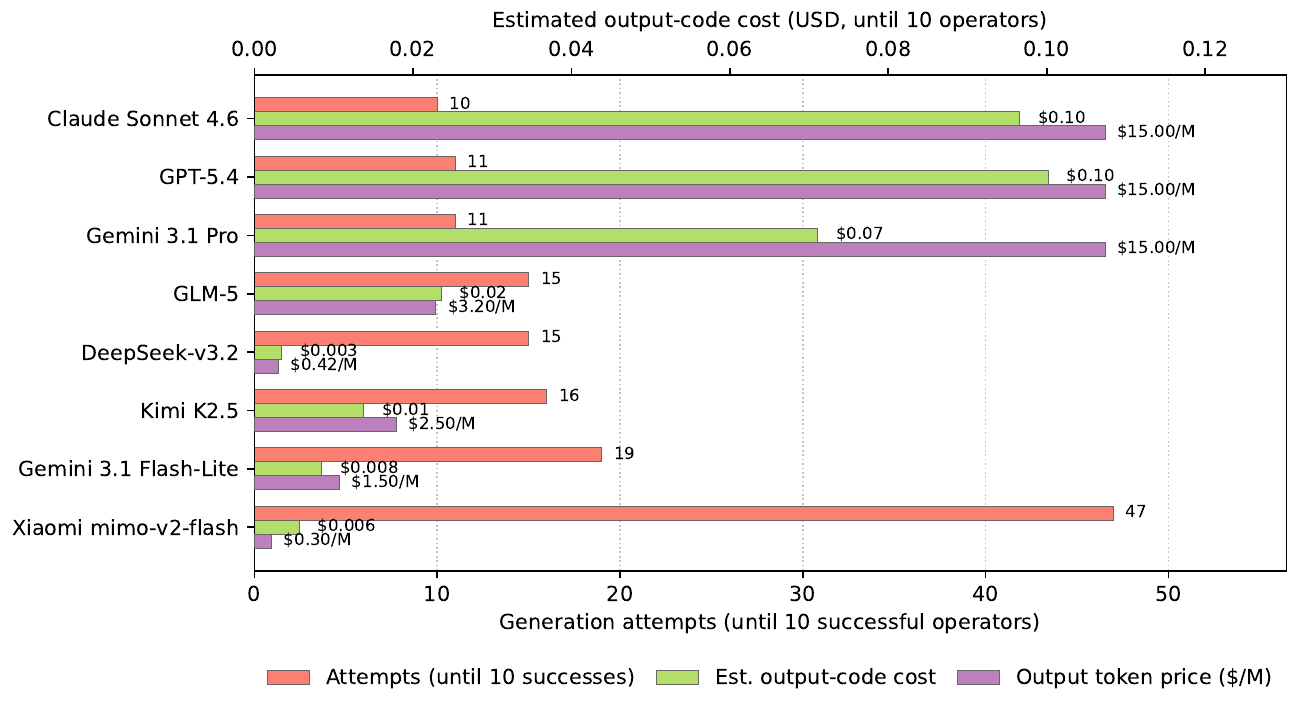}
        \caption{Generation attempts until ten valid operators, estimated cost of all generated code, including failed attempts, and price per million tokens.}
        \label{fig:zero-shot-attempts-and-output-price}
    \end{figure*}

    \paragraph{Synthesis effort.}
    \Cref{fig:zero-shot-attempts-and-output-price} summarizes synthesis cost and effort for each model. Xiaomi mimo-v2-flash has the lowest price per million tokens but requires the most attempts and remains weaker on training and test $R^2$ as shown in \Cref{tab:zero-shot-patterns,tab:zero-shot-test-summary}. Claude Sonnet~4.6, GPT-5.4, and Gemini~3.1 Pro come at higher listed prices but require fewer attempts and achieve stronger aggregate operator quality. GLM-5 and Kimi~K2.5 strike a cost-effective middle ground among price, synthesis effort, and empirical performance.

    \subsection{RQ2: Patterns in Generated Operators}\label{subsec:rq2-patterns}

    \begin{table}[!t]
        \centering
        \caption{Structural properties of each LLM's best-performing generated operator on the training set.}
        \label{tab:pattern-characteristics}
        \begingroup
        \small
        \setlength{\tabcolsep}{3pt}
        \resizebox{\columnwidth}{!}{%
            \begin{tabular}{lcccccc}
                \toprule
                Model                 & Stage-aware  & Difficulty-aware & Diversity signal & Rank transform & Selection mechanism & Elite injection \\
                \midrule
                Claude Sonnet 4.6     & $\checkmark$ &                  & $\checkmark$     &                & Tournament                  &                 \\
                Gemini 3.1 Flash-Lite & $\checkmark$ &                  &                  &                & Boltzmann                   &                 \\
                GLM-5                 & $\checkmark$ & $\checkmark$     & $\checkmark$     &                & Tournament                  &                 \\
                GPT-5.4               & $\checkmark$ &                  & $\checkmark$     & $\checkmark$   & Boltzmann selection + elite & $\checkmark$    \\
                Xiaomi mimo-v2-flash  & $\checkmark$ &                  &                  &                & Tournament                  &                 \\
                DeepSeek-v3.2         & $\checkmark$ &                  & $\checkmark$     &                & Tournament                  &                 \\
                Gemini 3.1 Pro        & $\checkmark$ & $\checkmark$     & $\checkmark$     & $\checkmark$   & Tournament                  &                 \\
                Kimi K2.5             & $\checkmark$ & $\checkmark$     & $\checkmark$     &                & Tournament                  &                 \\
                \bottomrule
            \end{tabular}%
        }
        \endgroup
    \end{table}

    \Cref{tab:pattern-characteristics} summarizes each model's training-best operator using six structural properties. Stage-aware scheduling denotes that selection pressure changes with evolutionary progress rather than staying fixed. Difficulty-aware weighting assigns different importance across fitness cases, such as curriculum schedules that emphasize harder or higher-variance cases. A diversity signal augments scalar training error with behavioral- or residual-based dissimilarity so that parents are not chosen from near-duplicate predictors alone. Rank transform indicates that within-population ranks replace raw fitness when individuals are compared. The selection mechanism names the parent-selection rule, covering tournaments, Boltzmann or roulette selection, or combinations of these. Elite injection reserves a fixed set for top-scoring individuals explicitly. In these best operators, stage-aware scheduling is universal. Many designs also pair fitness with either difficulty-aware case weighting or behavior-level diversity, so the strongest zero-shot rules rarely rely on a single global error scalar. By contrast, explicit rank transforms and elite injection appear only in a subset of operators. For the selection backbone, tournament-style mechanisms remain the most common, alongside a few Boltzmann-style variants.

    For illustration, \Cref{alg:pattern-kimi-k25} gives abbreviated pseudocode for a representative Kimi K2.5-generated selection operator that achieved the highest median training $R^2$. A representative GPT-5.4-generated selection operator that achieved the highest median test $R^2$, together with a brief discussion, is given in the appendix\footnote{Supplementary material is available at \url{https://github.com/hengzhe-zhang/ppsn2026-zero-shot-llm-selection/blob/main/supplementary_material.pdf}.} in \Cref{alg:pattern-gpt54}. Both use population size $n=|\mathcal{P}|$, requested number of parents~$k$, normalized evolutionary stage $t\in[0,1]$, and stability constant~$\varepsilon$.

    \begin{algorithm}[!t]
\caption{A representative Kimi K2.5-generated selection operator}
\label{alg:pattern-kimi-k25}
\begin{algorithmic}[1]
    \STATE \textbf{Require:} population $\mathcal{P}=\{p_i\}_{i=1}^n$, number of parents $k\in\mathbb{N}$, stage $t \in [0,1]$, stability constant $\varepsilon>0$, case-wise error matrix $R\in\mathbb{R}^{n\times m}$, column-normalized errors $R_{\mathrm{n}}$ per fitness case.
    \STATE $\mathbf{c}\leftarrow \frac{1}{n}R_{\mathrm{n}}^\top\mathbf{1}_n$ \COMMENT{mean error vector}
    \STATE $\eta\leftarrow \mathbf{1}_n-|R_{\mathrm{n}}\mathbf{c}|$ \COMMENT{residual-based novelty}
    \STATE $v_j,\quad j=1,\ldots,m$ \COMMENT{variance per fitness case}
    \STATE $\overline{v}\leftarrow \frac{1}{m}\sum_j v_j,\quad \mathbf{v}\leftarrow(v_1,\ldots,v_m)^\top$ \COMMENT{aggregate case statistics}
    \STATE Curriculum weights: emphasize high-variance cases as stage $t$ increases. \label{line:kimi-curriculum}\\[0.35em]
    $\mathbf{w}\leftarrow (1-t)\mathbf{1}_m+t\,\mathbf{v}/(\overline{v}+\varepsilon)$
    \STATE $f\leftarrow -\frac{1}{m}(R_{\mathrm{n}}\odot R_{\mathrm{n}})\mathbf{w}$ \COMMENT{weighted training error}
    \STATE $c\leftarrow 1-\mathrm{normalize}(|p|)$ \COMMENT{tree-size parsimony}
    \STATE Weighted mean selection score $s$: $z(\cdot)$ standardizes values within $\mathcal{P}$. \label{line:kimi-scalarization}\\[0.35em]
    $s\leftarrow t\cdot z(f)+(1-t)\cdot z(\eta)+0.1\,c$
    \STATE $\tau\leftarrow \max(2,\lfloor 2+6t\rfloor)$ \COMMENT{tournament size schedule}\label{line:kimi-tournament}
    \STATE $T\in\{1,\ldots,n\}^{k\times \tau}$ \COMMENT{random tournament draws}
    \FOR{$u = 1$ \textbf{to} $k$}
    \STATE $i^\star_u\leftarrow \arg\max_{v\in\{1,\ldots,\tau\}} s_{T_{uv}}$ \COMMENT{best index in each tournament}
    \ENDFOR
    \STATE $\mathcal{S}\leftarrow \{p_{T_{u,i^\star_u}}\}_{u=1}^k$ \COMMENT{parent mating pool}
    \STATE \textbf{return} $\mathcal{S}$ with $|\mathcal{S}|=k$.
\end{algorithmic}
    \end{algorithm}

    \paragraph{Kimi K2.5.}
    Intuitively, the key idea in this Kimi K2.5-designed operator is an adaptive curriculum over fitness cases, implemented through stage-dependent weights $w_j$. As $t$, the current generation divided by the total number of generations, increases, the operator assigns larger $w_j$ to cases with higher variance $v_j$, rather than treating all cases equally throughout evolution. Under this design, selection pressure is gradually redirected toward unresolved parts of the regression task where the current population still shows substantial disagreement.

    This behavior is realized through a weighted training-error score\linebreak
    $f_i=-m^{-1}\sum_j w_j e_{ij}^2$. In \Cref{alg:pattern-kimi-k25}, Line~\ref{line:kimi-curriculum} increases the weight on high-variance fitness cases as evolution proceeds, so selection focuses more on cases where the population still disagrees; Line~\ref{line:kimi-scalarization} combines this weighted error with novelty, so early search remains more exploratory; and Line~\ref{line:kimi-tournament} increases tournament size over time, thereby raising selection pressure in later generations.

    A prominent pattern is that Kimi's generated operators compare each individual's error vector in $R_{\mathrm{n}}\in\mathbb{R}^{n\times m}$ against a population-level aggregate, here the mean error vector $\mathbf{c}=\frac{1}{n}R_{\mathrm{n}}^\top\mathbf{1}_n$, rather than computing all pairwise dissimilarities among population members. The novelty term $\eta=\mathbf{1}_n-|R_{\mathrm{n}}\mathbf{c}|$ is then derived from this reference point. For population size $|\mathcal{P}|=n$ and $m$ fitness cases per evaluation, constructing and applying this aggregate reference requires $\mathcal{O}(nm)$ operations per selection call, whereas an explicit all-pairs dissimilarity pass scales as $\mathcal{O}(n^2m)$ when each pairwise comparison inspects all cases. Because the prompt does not explicitly guide the design of an efficient operator, this aggregate-based approach cannot be attributed to explicit domain knowledge encoded in the prompt text. This suggests that LLMs can design efficient algorithms even in zero-shot settings.

    \subsection{RQ3: Exploratory Alignment with General LLM Leaderboards}
    Algorithm-design tasks still receive limited coverage in mainstream LLM leaderboards. In this context, it is natural to ask which \emph{existing} public indicators might help identify suitable models before running expensive, domain-specific evaluations. We compare widely used benchmark scores with our GP-task summaries, with associations summarized using Spearman rank correlation~$\rho$. Several table cells are empty because not all LLMs report scores on every benchmark.

    \begin{table*}[!t]
        \centering
        \footnotesize
        \caption{Reference benchmark scores and rank correlations with zero-shot selection performance. Correlations greater than 0.9 are shown in bold.}
        \resizebox{\textwidth}{!}{%
            \begin{tabular}{l|c|c|c|c|c|c|c|c}
                \toprule
                Method                               & GPQA-Diamond & SWE-bench\ Verified & ARC-AGI\ v2 & MMMLU         & BrowseComp & MMMU-Pro & MCP\ Atlas & HLE \\
                \midrule
                Claude Sonnet 4.6                    & 89.9         & 79.6                & 58.3        & 89.3          & 74.7       & 75.6     & 61.3       & 49            \\
                GLM-5                                & --           & 77.8                & --          & --            & 75.9       & --       & 67.8       & --            \\
                Gemini 3.1 Flash-Lite                & 86.9         & --                  & --          & 88.9          & --         & 76.8     & --         & 16            \\
                Gemini 3.1 Pro                       & 94.3         & 80.6                & 77.1        & 92.6          & 85.9       & 80.5     & 69.2       & 51.4          \\
                GPT-5.4                              & 92.8         & --                  & 73.3        & --            & 82.7       & 81.2     & 67.2       & 39.8          \\
                Xiaomi mimo-v2-flash                 & 83.7         & 73.4                & --          & --            & 58.3       & --       & --         & 22.1          \\
                DeepSeek-v3.2                        & --           & --                  & --          & --            & --         & --       & --         & --            \\
                Kimi K2.5                            & 87.6         & 76.8                & --          & --            & 74.9       & 78.5     & --         & 50.2          \\
                \midrule
                Spearman $\rho$ vs.\ test mean       & 0.38         & 0.70                & -0.50       & 0.00          & 0.14       & -0.82    & -0.40      & 0.29          \\
                Spearman $\rho$ vs.\ test median     & 0.03         & 0.60                & -0.50       & -0.50         & -0.14      & -0.70    & -0.20      & -0.20         \\
                Spearman $\rho$ vs.\ test best       & 0.43         & 0.50                & 0.50        & 0.50          & 0.71       & 0.70     & 0.40       & -0.09         \\
                Spearman $\rho$ vs.\ training mean   & 0.66         & 0.90                & -0.50       & 0.50          & 0.26       & -0.30    & -0.20      & 0.83          \\
                Spearman $\rho$ vs.\ training median & 0.77         & \textbf{1.00}       & 0.50        & \textbf{1.00} & 0.49       & 0.00 & 0.40 & \textbf{0.94} \\
                Spearman $\rho$ vs.\ training best   & 0.20         & 0.30                & -0.50       & 0.50          & -0.14      & -0.20    & -0.20      & 0.77          \\
                \bottomrule
            \end{tabular}%
        }
        \label{tab:benchmark-reference}
    \end{table*}

    \paragraph{Training $R^2$.}
    The most striking pattern in \Cref{tab:benchmark-reference} is the strong Spearman $\rho$ between Humanity's Last Exam and training $R^2$, and between SWE-bench Verified and training $R^2$---for example, $\rho=0.94$ and $\rho=1.00$, respectively, against the training median---which makes them the most promising practical rank-order surrogates for training parent-selection operator quality. These values reflect correlation, not causation, but they may still guide practitioners when prioritizing models before costly domain-specific runs.

    \paragraph{Test $R^2$.}
    The link to held-out test $R^2$ is more modest. For example, the correlation between SWE-bench Verified and test median is only 0.6. This remains an open question. Overfitting and train--test shift interact with representation, fitness definition, and other parts of the GP pipeline, and improvements in parent selection alone cannot fully resolve overfitting issues, so leaderboard scores align less clearly with test $R^2$ in this sample.

    \section{Conclusions}\label{sec:conclusion}
    This paper presented a zero-shot benchmark for LLM-synthesized parent-selection operators in GP for symbolic regression. The experimental results show that Claude Sonnet~4.6 and Gemini~3.1 Pro stand out for consistently strong performance on both training and test $R^2$ across ten independent zero-shot operators, and that the strongest operator in our study---generated by Kimi~K2.5---exceeds the popular automatic lexicase baseline in search effectiveness, as evidenced by its stronger performance on the training set. The generated operators exhibit case-wise discrimination, for example by assigning higher weights to more discriminative instances, without explicit prompting beyond exposure to per-case errors. LLMs also demonstrate an ability to design efficient code. The rank correlations identify Humanity's Last Exam and SWE-bench Verified as promising off-the-shelf rank-order surrogates for training $R^2$ in algorithm design, while their relationship to held-out test $R^2$ remains unclear. One advantage of this benchmark over SWE-bench is that it naturally resists data contamination, since a strong selection operator must be genuinely novel rather than a memorized solution. For future work, it is worth exploring whether extended reasoning budgets improve operator quality, whether post-processing can yield simpler and more interpretable operators, and how to improve generalization, such as through LLM-designed regularization terms.

    \section*{Acknowledgements}
    This work was supported in part by the Marsden Fund of New Zealand Government under Contract VUW1913, and Contract VUW2016; in part by the MBIE Data Science SSIF Fund under Contract RTVU1914; in part by the MBIE Endeavor Research Programme under Contract UOCX2104; and in part by the Catalyst: Leaders International Leader Fellowship grant under Contract 23-VUW-006-ILF.

    \bibliographystyle{splncs04}
    \bibliography{zero_shot_refs}

    \clearpage
    \appendix

    This appendix provides supplementary material for \Cref{sec:results-analysis}.

    \begin{itemize}
        \item \Cref{sec:appendix-robustness} reports robustness results under alternative variation probabilities and supports the empirical findings in \Cref{subsec:rq1-zero-shot-performance}.
        \item \Cref{sec:appendix-gpt54} presents a representative GPT-5.4-generated selection operator and supports the pattern analysis in \Cref{subsec:rq2-patterns}.
    \end{itemize}

    \section{Robustness to Different Variation Probabilities}\label{sec:appendix-robustness}

    As a robustness check, \Cref{tab:zero-shot-training-summary-08,tab:zero-shot-test-summary-08} report the training- and test-set results from a supplementary experiment with crossover probability $0.8$ and mutation probability $0.2$. The results are consistent with the main experiment. On training, Claude Sonnet~4.6 and Gemini~3.1 Pro remain the strongest models in the aggregate summaries, while Kimi~K2.5 still yields the strongest single training operator according to the best-score column. On test, Claude Sonnet~4.6 and Gemini~3.1 Pro again remain among the top-performing models, with only modest shifts in rankings across the summary metrics.

    \begin{table}[H]
        \centering
        \caption{Zero-shot \textbf{training} performance summary across benchmarked LLMs with $N=200$, $G=100$, crossover probability $0.8$, and mutation probability $0.2$. Parenthetical integers rank models within each metric column.}
        \resizebox{\columnwidth}{!}{%
            \begin{tabular}{c|c|c|c|c|c|c}
                \toprule
                Model                 & Mean over mean      & Median over mean    & Mean over median    & Median over median  & Best score & Mean rank \\
                \midrule
                Claude Sonnet 4.6     & \textbf{0.7025 (1)} & \textbf{0.6946 (1)} & \textbf{0.7972 (1)} & \textbf{0.7946 (1)} & 0.8418 (5) & \textbf{1.80} \\
                Gemini 3.1 Pro        & 0.6981 (2)          & 0.6930 (2)          & 0.7939 (2)          & 0.7893 (2)          & 0.8504 (4)          & 2.40          \\
                GLM-5                 & 0.6889 (3)          & 0.6893 (3)          & 0.7711 (3)          & 0.7763 (3)          & 0.8510 (3)          & 3.00          \\
                Kimi K2.5             & 0.6868 (4)          & 0.6889 (4)          & 0.7598 (4)          & 0.7670 (4)          & \textbf{0.8776 (1)} & 3.40          \\
                GPT-5.4               & 0.6646 (6)          & 0.6793 (5)          & 0.7111 (5)          & 0.7418 (6)          & 0.8557 (2)          & 4.80          \\
                Xiaomi mimo-v2-flash  & 0.6657 (5)          & 0.6739 (6)          & 0.6993 (6)          & 0.7519 (5)          & 0.8009 (6)          & 5.60          \\
                Gemini 3.1 Flash-Lite & 0.6539 (7)          & 0.6670 (7)          & 0.6970 (7)          & 0.7382 (7)          & 0.7892 (7)          & 7.00          \\
                DeepSeek-v3.2         & 0.6211 (8)          & 0.6271 (8)          & 0.6156 (8)          & 0.5753 (8)          & 0.7664 (8)          & 8.00          \\
                \bottomrule
            \end{tabular}
        }
        \label{tab:zero-shot-training-summary-08}
    \end{table}

    \begin{table}[H]
        \centering
        \caption{Zero-shot \textbf{test} performance summary across benchmarked LLMs with $N=200$, $G=100$, crossover probability $0.8$, and mutation probability $0.2$. Parenthetical integers rank models within each metric column.}
        \resizebox{\columnwidth}{!}{%
            \begin{tabular}{c|c|c|c|c|c|c}
                \toprule
                Model                 & Mean over mean      & Median over mean    & Mean over median    & Median over median  & Best score & Mean rank \\
                \midrule
                Claude Sonnet 4.6     & 0.5439 (2)          & \textbf{0.5609 (1)} & \textbf{0.5938 (1)} & 0.5832 (2)          & 0.6419 (3) & \textbf{1.80} \\
                Gemini 3.1 Pro        & \textbf{0.5535 (1)} & 0.5514 (2)          & 0.5899 (2)          & 0.5823 (3)          & \textbf{0.6454 (1)} & \textbf{1.80} \\
                Kimi K2.5             & 0.5306 (5)          & 0.5402 (6)          & 0.5836 (3)          & \textbf{0.5889 (1)} & 0.6440 (2)          & 3.40          \\
                GLM-5                 & 0.5439 (2)          & 0.5471 (3)          & 0.5803 (4)          & 0.5784 (4)          & 0.6017 (6)          & 3.80          \\
                Gemini 3.1 Flash-Lite & 0.5350 (4)          & 0.5468 (4)          & 0.5634 (5)          & 0.5680 (7)          & 0.6108 (5)          & 5.00          \\
                GPT-5.4               & 0.5280 (6)          & 0.5355 (7)          & 0.5590 (6)          & 0.5770 (6)          & 0.6174 (4)          & 5.80          \\
                Xiaomi mimo-v2-flash  & 0.5227 (7)          & 0.5439 (5)          & 0.5488 (7)          & 0.5781 (5)          & 0.5975 (7)          & 6.20          \\
                DeepSeek-v3.2         & 0.4678 (8)          & 0.5057 (8)          & 0.5100 (8)          & 0.5262 (8)          & 0.5899 (8)          & 8.00          \\
                \bottomrule
            \end{tabular}
        }
        \label{tab:zero-shot-test-summary-08}
    \end{table}

    \section{Representative GPT-5.4-Generated Selection Operator}\label{sec:appendix-gpt54}
    \begin{algorithm}[!t]
\caption{A representative GPT-5.4-generated selection operator}
\label{alg:pattern-gpt54}
\begin{algorithmic}[1]
    \STATE \textbf{Require:} population $\mathcal{P}=\{p_i\}_{i=1}^n$, number of parents $k\in\mathbb{N}$, stage $t \in [0,1]$, stability constant $\varepsilon>0$, prediction matrix $P\in\mathbb{R}^{n\times d}$, per-individual mean training errors $\bar{e}=(\bar{e}_1,\ldots,\bar{e}_n)^\top$, syntax-tree depths $h=(h_1,\ldots,h_n)^\top$ with $h_i=\mathrm{height}(p_i)$, program sizes $|p|=(|p_1|,\ldots,|p_n|)^\top$, e.g.\ node counts.
    \STATE $S\leftarrow \mathrm{clip}(PP^\top,[-1,1])$ \COMMENT{clipped similarity matrix}
    \STATE $d\leftarrow \mathbf{1}_n-\frac{1}{n}S\mathbf{1}_n$ \COMMENT{prediction-space dispersion}
    \STATE Min--max normalize within $\mathcal{P}$, for $i=1,\ldots,n$: \\[0.35em]
    $\displaystyle q_i^{\mathrm{fit}} \leftarrow 1-\frac{\bar{e}_i-\min_{\ell}\bar{e}_\ell}{\max_{\ell}\bar{e}_\ell-\min_{\ell}\bar{e}_\ell+\varepsilon},\quad
    q_i^{\mathrm{div}} \leftarrow \frac{d_i-\min_{\ell}d_\ell}{\max_{\ell}d_\ell-\min_{\ell}d_\ell+\varepsilon}$ \\[0.35em]
    $\displaystyle q_i^{\mathrm{sz}} \leftarrow 1-\frac{|p|_i-\min_{\ell}|p|_\ell}{\max_{\ell}|p|_\ell-\min_{\ell}|p|_\ell+\varepsilon},\quad
    q_i^{\mathrm{ht}} \leftarrow 1-\frac{h_i-\min_{\ell}h_\ell}{\max_{\ell}h_\ell-\min_{\ell}h_\ell+\varepsilon}$
    \STATE $\nu \leftarrow 0.65(1-t)+0.15$ \COMMENT{diversity blend weight}\label{line:gpt-diversity}
    \STATE $\beta \leftarrow 0.15+0.55t$ \COMMENT{parsimony blend weight}
    \STATE Composite score: fit, diversity, and parsimony, weighted by evolutionary stage. \label{line:gpt-composite}\\[0.35em]
    $q \leftarrow (1.2+1.8t)\,q^{\mathrm{fit}}+\nu\,q^{\mathrm{div}}+\tfrac{\beta}{2}(q^{\mathrm{sz}}+q^{\mathrm{ht}})$
    \STATE $m_e\leftarrow \min\bigl(n,\lceil k(0.08+0.12t)\rceil\bigr)$ \COMMENT{number of elite slots}\label{line:gpt-elite}
    \STATE $\mathcal{E}\leftarrow \bigl\{\text{top } m_e \text{ indices by } q\bigr\}$ \COMMENT{elite parent indices}
    \STATE $\theta \leftarrow \max(0.05,\,0.9-0.7t)$ \COMMENT{softmax temperature}
    \STATE $\ell\leftarrow q/\theta-\max(q)/\theta$ \COMMENT{shifted selection logits}
    \STATE $\delta \leftarrow 0.92-0.25(1-t)$ \COMMENT{similarity crowding threshold}
    \STATE Crowding-adjusted weights and softmax mass for selection. \label{line:gpt-crowding}\\[0.35em]
    $g\leftarrow \bigl(\mathbf{1}+(S>\delta)\mathbf{1}_n\bigr)^{-1}$ \\[0.35em]
    $m\leftarrow \exp(\ell)\odot(0.35+0.65 g)$ \COMMENT{unnormalized softmax weights}
    \STATE For each elite index $i\in\mathcal{E}$, down-weight $m_i$.
    \STATE $\omega\leftarrow m/\sum_j m_j$ \COMMENT{softmax selection probabilities}
    \STATE Sample non-elite parents i.i.d.\ from $\mathrm{Cat}(\omega)$, then merge with elites into~$\mathcal{S}$.
    \STATE \textbf{return} $\mathcal{S}$ with $|\mathcal{S}|=k$.
\end{algorithmic}
    \end{algorithm}

    In \Cref{alg:pattern-gpt54}, the key idea of the GPT-5.4-designed operator is phenotypic diversity control. Rather than distinguishing individuals only by their mean training error $\bar{e}_i$, it uses their semantics, represented by output vectors in the prediction matrix $P$, to identify phenotypic duplicates and de-emphasize them during parent selection. Concretely, the operator steers the mating pool toward behaviorally diverse individuals early in the run when stage $t$ is small, while still allowing stronger emphasis on fitness in later generations.

    This behavior is realized through the similarity matrix $S=PP^\top$ and the derived dispersion score $d_i$. In \Cref{alg:pattern-gpt54}, Line~\ref{line:gpt-composite} combines fitness, diversity, and parsimony into a stage-dependent score, with the diversity weight $\nu$ in Line~\ref{line:gpt-diversity} decreasing as $t$ grows so that search gradually shifts from exploration toward exploitation. Line~\ref{line:gpt-elite} preserves a small elite block of high-scoring parents, while Line~\ref{line:gpt-crowding} down-weights individuals in high-similarity neighborhoods before Boltzmann selection. This hybrid design keeps strong parents but avoids filling the mating pool with semantically similar programs.

\end{document}